\documentclass[final,5p,times,twocolumn]{elsarticle}

\usepackage{lineno,hyperref}

\usepackage{amsmath}
\usepackage{amssymb}
\usepackage{color}
\usepackage{multirow}
\usepackage{array}
\usepackage{epstopdf}
\usepackage{subfigure}
\usepackage{enumitem}
\usepackage{booktabs}
\usepackage{verbatim}
\biboptions{sort&compress}

\begin{document}

\begin{frontmatter}

\title{LC$^3$Net: Ladder context correlation complementary network for salient object detection}

\author[Address1]{Xian Fang}
\ead{xianfang@mail.nankai.edu.cn}
\author[Address2]{Jinchao Zhu}
\author[Address1]{Xiuli Shao}
\author[Address2]{Hongpeng Wang\corref{*}}
\cortext[*]{Corresponding author.}
\ead{hpwang@nankai.edu.cn}
\address[Address1]{College of Computer Science, Nankai University, Tianjin 300350, China}
\address[Address2]{College of Artificial Intelligence, Nankai University, Tianjin 300350, China}

\begin{abstract}
Currently, existing salient object detection methods based on convolutional neural networks commonly resort to constructing discriminative networks to aggregate high level and low level features. However, contextual information is always not fully and reasonably utilized, which usually causes either the absence of useful features or contamination of redundant features. To address these issues, we propose a novel ladder context correlation complementary network (LC$^3$Net) in this paper, which is equipped with three crucial components. At the beginning, we propose a filterable convolution block (FCB) to assist the automatic collection of information on the diversity of initial features, and it is simple yet practical. Besides, we propose a dense cross module (DCM) to facilitate the intimate aggregation of different levels of features by validly integrating semantic information and detailed information of both adjacent and non-adjacent layers. Furthermore, we propose a bidirectional compression decoder (BCD) to help the progressive shrinkage of multi-scale features from coarse to fine by leveraging multiple pairs of alternating top-down and bottom-up feature interaction flows. Extensive experiments demonstrate the superiority of our method against 16 state-of-the-art methods.
\end{abstract}


\end{frontmatter}

\section{Introduction}

Salient object detection, also known as saliency detection, is derived with the goal of detecting the most conspicuous and distinctive regions in the scene. Due to the capability and scalability, it has been widely applied in computer vision tasks, such as image parsing \cite{Lai2016Saliency}, image captioning \cite{Zhou2019Re-caption}, image segmentation \cite{Chang2011From, Qin2014Integration}, image retrieval \cite{Gao2015Database, Wang2020Visual}, visual tracking \cite{Zhang2009Visual, Mahadevan2012Biologically, Hong2015Online} and many more.

Recently, benefiting from the explosive development of convolutional neural networks (CNNs), a lot of CNNs-based saliency detection methods are constantly emerging, and encouraging progresses have been made \cite{Wu2019A, Feng2020Residual, Feng2020CACNet, Zhao2020Suppress, Chen2020Global, Wang2020Progressive, Zhou2020Interactive, Wei2020Label, Luo2021LF^3Net}. Such methods conform to the encoder-decoder pattern, in which the hierarchical features of different levels are first extracted from the encoder and then propagated to the decoder to predict saliency maps. Many existing methods commonly resort to strengthening the feature representation via investigating components of the model, thereby constructing discriminative networks to aggregate high level features and low level features.

\begin{figure}
  \centering
  \includegraphics[width=0.4727\textwidth]{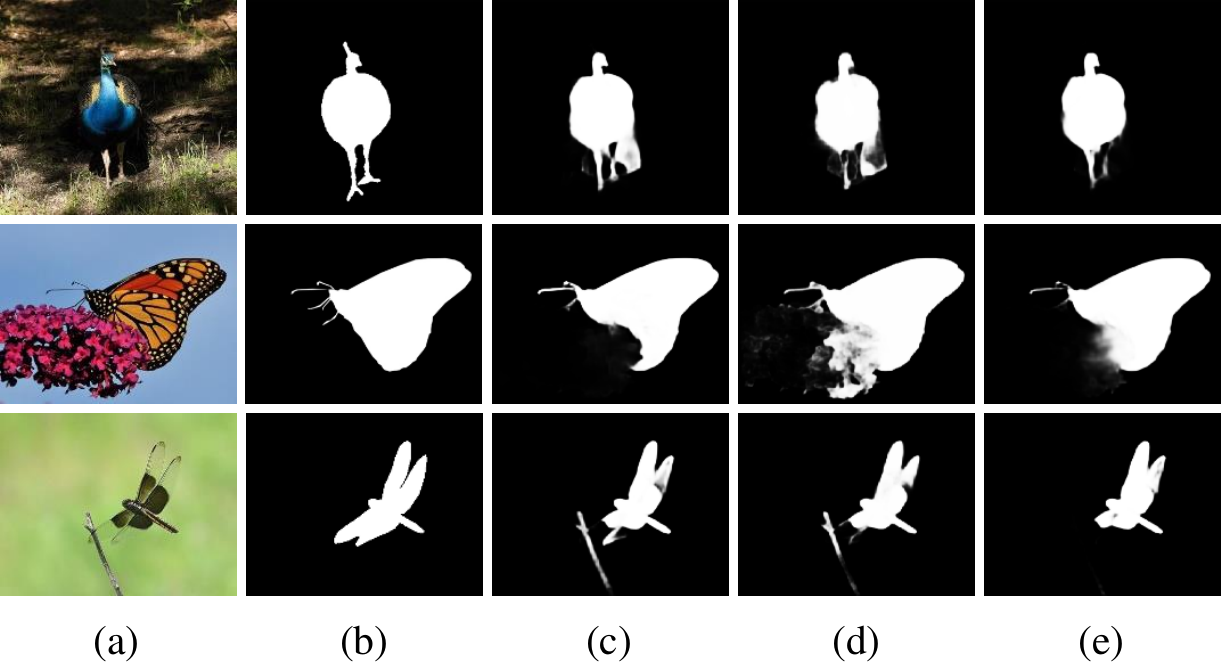}
  \caption{Some saliency maps obtained by different methods. (a) Image; (b) Ground truth; (c) F$^3$Net \cite{Wei2020F^3Net}; (d) MINet \cite{Pang2020Multi-scale}; (e) Ours.}
  %\vspace{-12pt}
  \label{Figure 1}
\end{figure}

As is known to all, high level features carry rich semantic information that possesses the relatively clear background, while low level features preserve enough detailed information that possesses the relatively clear boundary. Because of the different receptive fields at different layers, there is a big gap between high level features and low level features. As a result, it remains dilemmas in the utilization of contextual information. Actually, whether a region is salient or not is depend on the context. In other words, the key of saliency detection lies in how to grasp the complementarity of contextual information harmoniously. However, the correlation of context is quite difficult to be fully and reasonably exploited, which will inevitably aggravate the misjudgment of detection. On the one hand, when the contextual information could not be fully utilized, it will lead to the lack of useful features. In this case, the detection of objects fails in tangibly identifying the contours. On the other hand, when the contextual information could not be reasonably utilized, it will bring about the pollution of redundant features. In this case, the detection of objects is prone to suffer from the interference of noises.

To tackle the aforementioned problems in the previously existing methods, we propose a novel network called ladder context correlation complementary network (LC$^3$Net), which explicitly explores the reliable inherent complementarity of correlated contextual information. Specifically, LC$^3$Net is equipped with three crucial components, which are filterable convolution block (FCB), dense cross module (DCM) and bidirectional compression decoder (BCD). At the beginning, FCB is plug and play, which is placed immediately behind the backbone network to assist the automatic collection of information on the diversity of initial features. Besides, DCM has two sub-components according to the two sampling types, the one is that with upsampling (DCM-U), and the other is that with downsampling (DCM-D). The two own the opposite and symmetrical structure, which fuses the features of each level with all other higher or lower levels with semantic information and detailed information of both adjacent and non-adjacent layers. These two modes are combined to facilitate the intimate aggregation of different levels of features. Furthermore, BCD has three distinguished sub-components, namely, BCD$_1$, BCD$_2$ and BCD$_3$. Each of them has a pair of alternating interaction flows on the pathway from top to down and the pathway from bottom to top in turn. These three parts are merged to complete the interaction of features to help the progressive shrinkage of multi-scale features from coarse to fine. Without bells and whistles, we report extensive experimental results on five datasets in terms of four evaluation metrics to demonstrate the superiority of our method against 16 state-of-the-art methods. As shown in Figure \ref{Figure 1}, the saliency maps obtained by our method can better suppress the surrounding background noises and display the expected boundary contours in comparison with those obtained by other methods.
%We also the impact of each component and sub-component in our method.
%To show the intuitively, we visualize the features of different levels in Figure \ref{Figure 2}.

%\begin{figure}
%  \centering
%  \includegraphics[width=0.5\textwidth]{2.pdf}
%  \caption{Some feature maps of different layers obtained by our methods. (a) Image; (b) Ground truth; (c) f$_5$; (d) %  f$_4$; (e) f$_3$; (f) f$_2$.}
%  \vspace{-12pt}
%  \label{Figure 2}
%\end{figure}

\begin{figure*} [ht]
  \centering
  \includegraphics[width=1\textwidth]{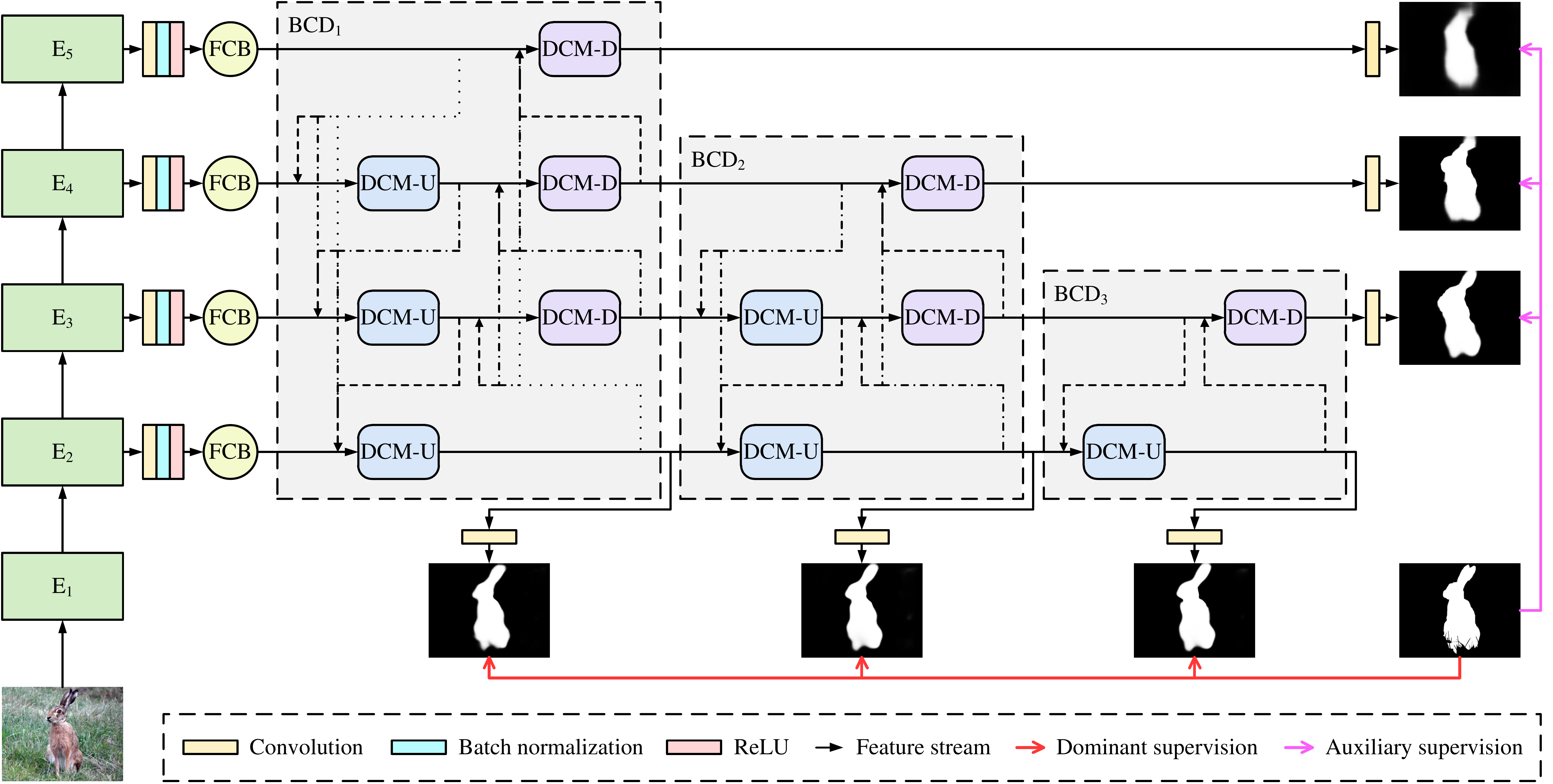}
  \caption{The overall architecture of the proposed ladder context correlation complementary network (LC$^3$Net). The filterable convolution block (FCB) is placed to dynamically collect the diversity information of initial features. The two modes of dense cross module (DCM), that with upsampling (DCM-U) and that with downsampling (DCM-D), are combined to tightly aggregate different levels of features. The three parts of bidirectional compression decoder (BCD), BCD$_{1}$, BCD$_{2}$ and BCD$_{3}$, are merged to gradually shrink the multi-scale features until refinement.}
  %\vspace{-12pt}
  \label{Figure 2}
\end{figure*}

The main contributions of this paper are summarized as follows:
\begin{itemize}
  \item A filterable convolution block (FCB) is proposed to dynamically collect the diversity information of initial features, and it is simple yet practical.
  \item A dense cross module (DCM) is proposed to tightly aggregate different levels of features by validly integrating semantic information and detailed information of both adjacent and non-adjacent layers.
  \item A bidirectional compression decoder (BCD) is proposed to gradually shrink the multi-scale features until refinement by leveraging multiple pairs of alternating top-down and bottom-up feature interaction flows.
  \item Experiments on five datasets in terms of four evaluation metrics prove that our proposed ladder context correlation complementary network (LC$^3$Net) achieves remarkable results compared with 16 state-of-the-art methods.
\end{itemize}

The remainder of this paper is organized as follows. In Section \ref{Section 2}, a brief review of the related works is stated. Section \ref{Section 3} describes the proposed method in detail. Section \ref{Section 4} analyzes the experimental results. Finally, the conclusion is given in Section \ref{Section 5}.

\section{Related Work}
\label{Section 2}

Throughout the past decades, numerous valuable salient object detection methods based on convolutional neural networks have been proposed. These methods are devoted to aggregate high level and low level features from major perspectives of short connection, attention mechanism, residual learning, gate unit and edge-aware knowledge.

%Amulet
\cite{Zhang2017Amulet} proposed to bridge all level features into the transport layer.
%DSS
\cite{Hou2017Deeply} proposed to introduce short connections into the skip-layer for saliency prediction.
%RAS
\cite{Chen2018Reverse} proposed to assemble reverse attention to serve side-output residual learning.
%C2S
\cite{Li2018Contour} proposed to borrow contour knowledge to predict the saliency maps.
%PAGR
\cite{Zhang2018Progressive} proposed to transfer information from deep layers to shallow layers by recurrent connection.
%PiCANet
\cite{Liu2018PiCANet} proposed to selectively attend to global or local pixel-wise contexts.
%BANet
\cite{Su2019Selectivity} proposed to revisit feature selectivity at boundaries and feature invariance at interiors simultaneously.
%EGNet
\cite{Zhao2019EGNet} proposed to guide the saliency detection with edge information.
%CPD
\cite{Wu2019Cascaded} proposed to abandon low level features in the cascaded framework.
%PoolNet
\cite{Liu2019A} proposed to tap the potentials of pooling and add edge information.
%GateNet
\cite{Zhao2020Suppress} proposed to adaptively control the amount of information flowing from each encoder to each decoder.
%F$^3$Net
\cite{Wei2020F^3Net} proposed to iteratively feedback features of different levels.
%MINet
\cite{Pang2020Multi-scale} proposed to collaboratively learn knowledge guidance through mutual learning and self-learning.

Although these above methods have exhibited promising prospects, there are still bottlenecks of dealing with the complex detection situations, which has plenty of space in the utilization of contextual information.

\section{Methodology}
\label{Section 3}

\subsection{Network Overview}

Figure \ref{Figure 2} shows the overall architecture of the proposed ladder context correlation complementary network (LC$^3$Net). For given backbone network, features $f_i$ $(i=1,2,3,4,5)$ from different layers can be easily extracted by the encoder \{E$_1$, E$_2$, E$_3$, E$_4$, E$_5$\}. Owing to the fact that feature $f_1$ imposes considerable computation cost and the performance improvement is negligible, we only focus on features $f_2$, $f_3$, $f_4$ and $f_5$ with the channels of \{256, 512, 1024, 2048\}. Through the combination of convolution, batch normalization and ReLU operations, these features are uniformly squeezed to 64 channels and separately transmitted to the block FCB. Next, the features are successively propagated to the special decoders BCD$_1$, BCD$_2$ and BCD$_3$ regularly stacked with multiple modules DCM-U and DCM-D. After that, the 3$\times$3 convolution is conducted to change the channel number to 1 for calculating the saliency scores.

\subsection{Filterable Convolution Block}

We propose the filterable convolution block (FCB) to diversely collect the diversity information of initial features produced from the backbone network.

\begin{figure} [t]
  \centering
  \includegraphics[width=0.475\textwidth]{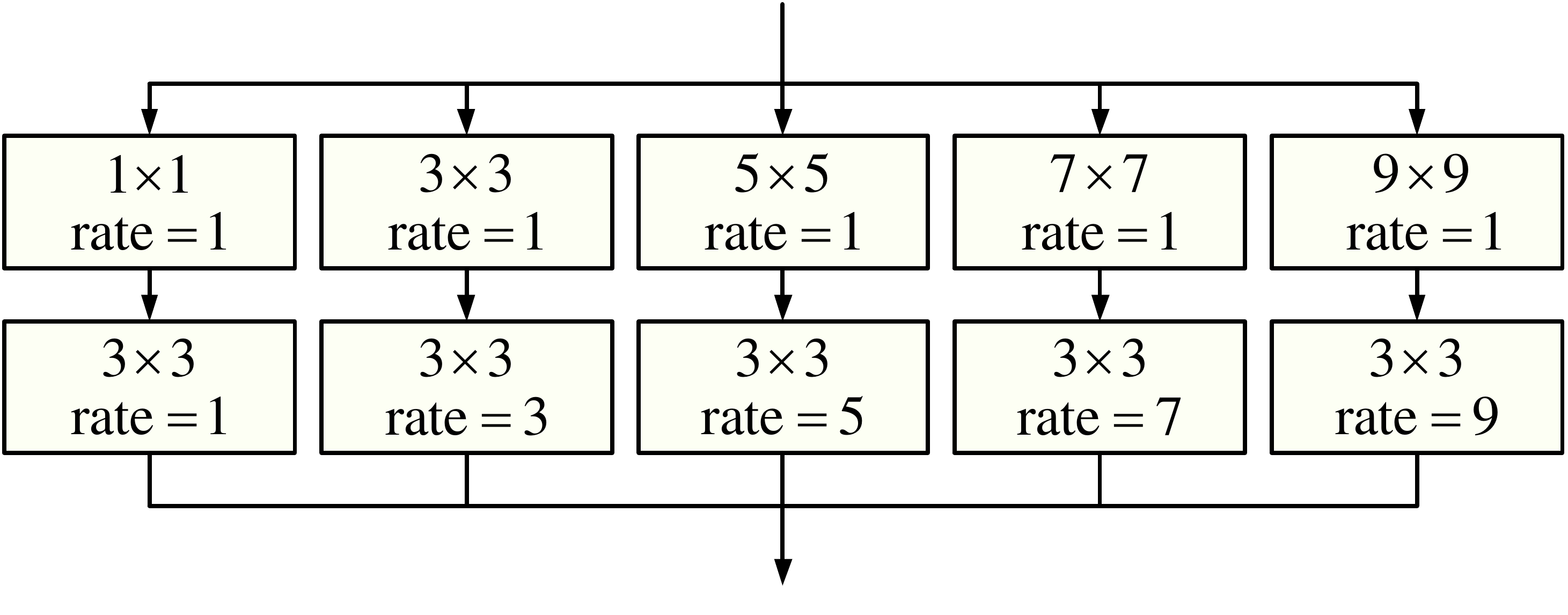}
  \caption{Core system of the proposed filterable convolution block (FCB).}
  %\vspace{-12pt}
  \label{Figure 3}
\end{figure}

There are five independent individual branches organized in FCB, which are connected in parallel. The core system is shown in Figure \ref{Figure 3}. Vertically, each branch is assigned a dilated convolution with different dilation rates. Horizontally, all of these branches have two steps. In the first step, all branches share the same dilation rate of 1, but the kernel sizes of those are 1$\times$1, 3$\times$3, 5$\times$5, 7$\times$7 and 9$\times$9, respectively. In the second step, all branches share the same kernel size of 3$\times$3, but the dilation rates of those are 1, 3, 5, 7 and 9, respectively.

At the end of the two steps, the collected information will be further concatenated and sent to a combination of convolution, batch normalization and ReLU operations.

\subsection{Dense Cross Module}

We propose the dense cross module (DCM) to tightly aggregate different levels of features, which powerfully promotes the integration of semantic information and detail information of both adjacent and non-adjacent layers, so as to maximize the mining of useful feature.

\begin{figure} [t]
  \centering
  \includegraphics[width=0.475\textwidth]{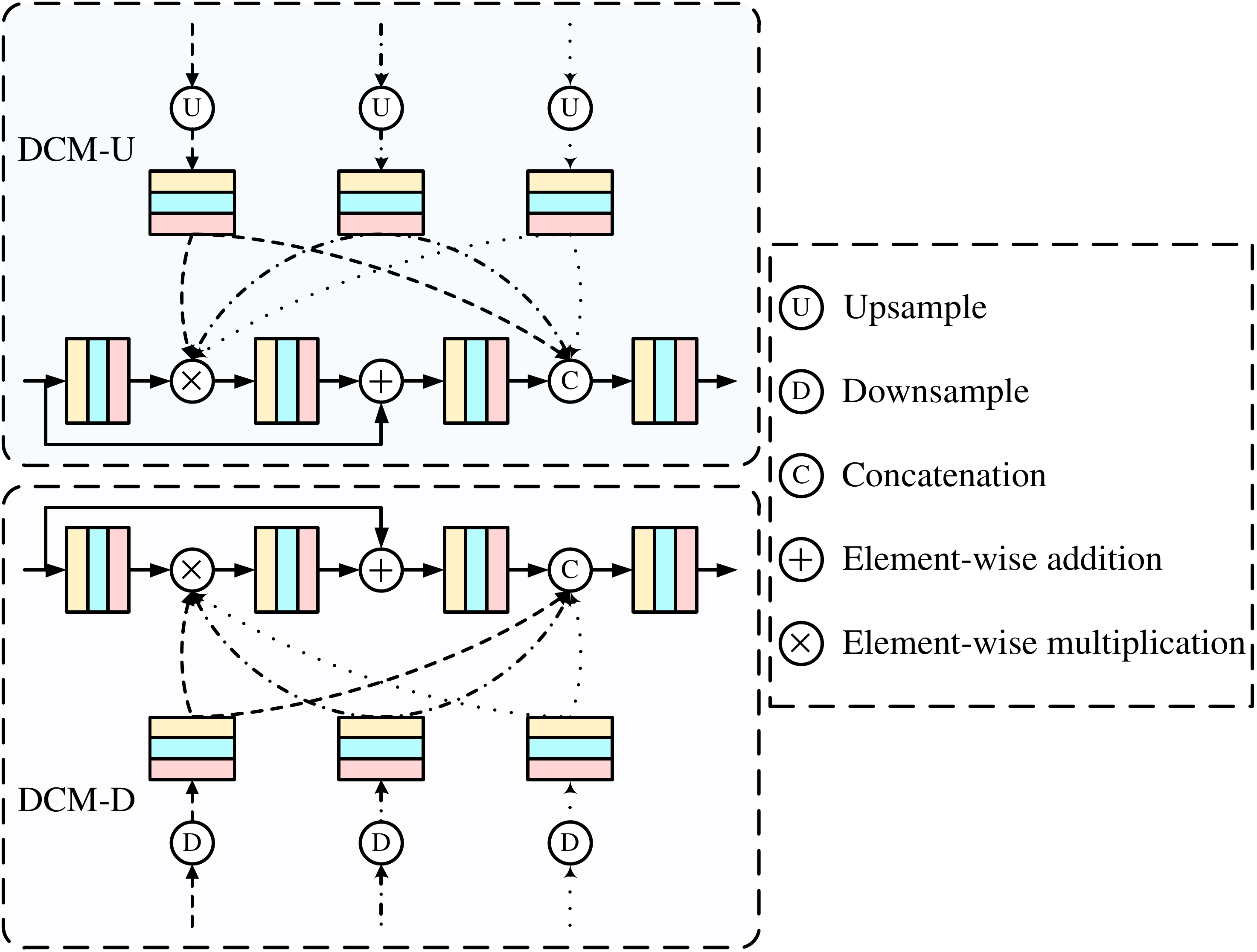}
  \caption{Illustrate of two modes of the proposed dense cross module (DCM).}
  %\vspace{-12pt}
  \label{Figure 4}
\end{figure}

DCM consists of two modes according to the two sampling types, that with upsampling (DCM-U) and that with downsampling (DCM-D), as shown in Figure \ref{Figure 4}. In particular, the former is committed to delivering high level features containing semantics to low level ones as much as possible, while the latter is committed to returning low level features containing details to high level ones as much as possible.

From the form of structure, DCM-U and DCM-D are just opposite and symmetrical. In both of them, all the features of higher levels or lower levels are upsampled or downsampled to fuse with the feature of each level. Such fusions go through an element-wise multiplication, an element-wise addition and a concatenation operation, as well as a series of combination of convolution, batch normalization and ReLU operations.

The specific process of DCM-U and DCM-D can be formulated as:
\begin{equation}
  \widehat{{f}_i} = \cup_{i}^{4} (\Gamma(Up(f_{i+1}))), i=4,3,2,
\end{equation}
\begin{equation}
  f'_i = \Gamma(Concat(\Gamma(\Gamma(\Gamma({f}_i) \otimes \widehat{{f}_i}) \oplus {f}_i), \widehat{{f}_i})), i=4,3,2,
\end{equation}
and
\begin{equation}
  \widehat{{f}_i} = \cup_{2}^{i-1} (\Gamma(Down(f_{i}))), i=3,4,5,
\end{equation}
\begin{equation}
  f'_i = \Gamma(Concat(\Gamma(\Gamma(\Gamma({f}_i) \otimes \widehat{{f}_i}) \oplus {f}_i), \widehat{{f}_i})), i=3,4,5,
\end{equation}
where $f_i$, $f'_i$ and $\widehat{{f}_i}$ $(i=2,3,4,5)$ represent the input feature, output feature and feature set of certain features after upsampling or downsampling of $i^{th}$ layer, respectively, $Up$, $Down$ and $Concat$ denote the upsample, downsample and concatenation operations, respectively, $\Gamma$ is the combination of convolution, batch normalization and ReLU operations, $\otimes$ denotes element-wise multiplication and $\oplus$ denotes element-wise addition.

\subsection{Bidirectional Compression Decoder}

We propose the bidirectional compression decoder (BCD) to shrink multi-scale features from coarse to fine in a progressive way, which is essentially a triple decoder and is built upon DCM.

As for the features to be concerned, they are thinly divided into three groups, \{$f_2$, $f_3$, $f_4$, $f_5$\}, \{$f_2$, $f_3$, $f_4$\} and \{$f_2$, $f_3$\}, which are involved in three important parts of BCD, namely, BCD$_1$, BCD$_2$ and BCD$_3$. In each part, DCM-U and DCM-D appear in pairs and excavate feature relying on the top-down and bottom-up manner in turn.

\begin{figure} [t]
  \centering
  \includegraphics[width=0.32\textwidth]{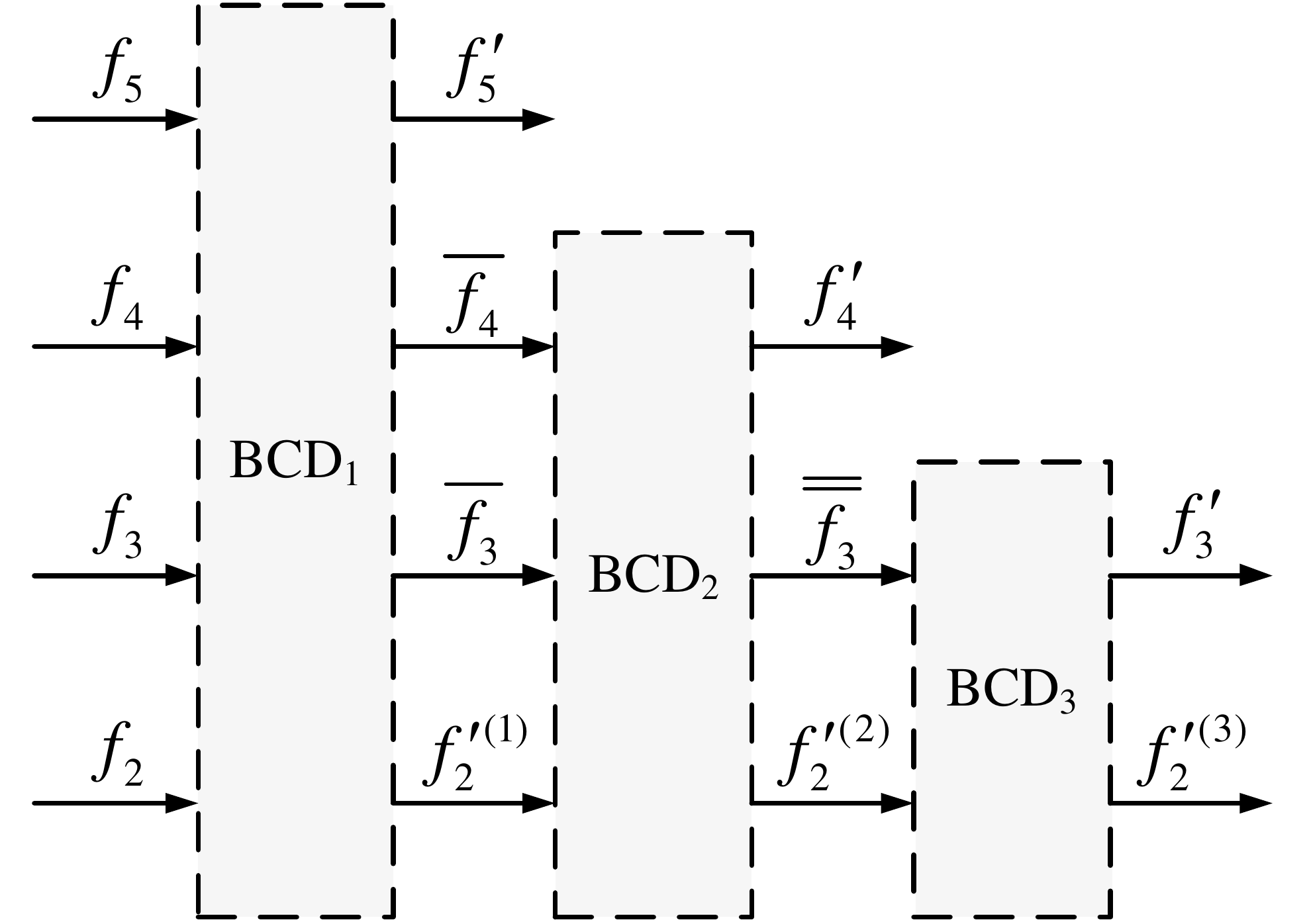}
  \caption{Interactive flow chart of three parts of the proposed bidirectional compression decoder (BCD).}
  %\vspace{-12pt}
  \label{Figure 5}
\end{figure}

Taking BCD$_1$ as an example, $f_4$ is updated after the original $f_{4}$ and $f_{5}$ pass through DCM-U for the first time, and $f_3$ is then updated after the original $f_3$ and $f_5$ along with the new $f_4$ pass through DCM-U for the second time, and $f_2$ continues to be updated after the original $f_2$ and $f_5$ along with the new $f_4$ and $f_3$ pass through DCM-U for the third time. So far, half of the procedure is completed on the pathway from top to down, the other half of the procedure on the pathway from bottom to up is exactly reversed. A similar set of procedures also act on BCD$_2$ and BCD$_3$. The interactive relationship between the three parts is shown in Figure \ref{Figure 5}, which generally looks like a three-tier ladder. It is worth pointing out that the design of three distinct parts instead of three identical parts can reduce the interference of redundant features on the premise of ensuring the sufficient interaction of features.

The rough process of feature interaction in BCD$_1$, BCD$_2$ and BCD$_3$ can be formulated as:
\begin{equation}
  \begin{split}
    (f'^{(1)}_2, \overline{f_3}, \overline{f_4}, f'_5)& = {\rm BCD_1}(f_5, f_4, f_3, f_2) \\
    (f'^{(2)}_2, \overline{\overline{f_3}}, f'_4)& = {\rm BCD_2}(\overline{f_4}, \overline{f_3}, f'^{(1)}_2), \\
    (f'^{(3)}_2, f'_3)& = {\rm BCD_3}(\overline{\overline{f_3}}, f'^{(2)}_2) \\
  \end{split}
\end{equation}
where $f_i$, $f'_i$ $(i=2,3,4,5)$, $\overline{f_i}$ $(i=3,4)$ and $\overline{\overline{f_i}}$ $(i=3)$ represent the input feature, output feature, first interaction intermediate feature and second interaction intermediate feature of $i^{th}$ layer, respectively.

Once the input features are acquired, after discarding all intermediate features, here three output features of $i^{nd}$ layer in the three parts will be obtained, which are regarded as the dominant stream features, and three output features of $5^{th}$, $4^{th}$ and $i^{rd}$ layers will be obtained, which are regarded as the auxiliary stream features.

\begin{table*} [t]
  \centering
  \caption{Quantitative comparison of different methods on five datasets in terms of four evaluation metrics. $\uparrow$ indicates that larger is better and $\downarrow$ indicates smaller is better. The best results are marked in \textbf{bold}. '--' means the authors have not provided corresponding saliency maps.}
  \renewcommand\tabcolsep{1.2pt}
  \renewcommand\arraystretch{1.1}
  \small
  \begin{tabular}{c|cccc|cccc|cccc|cccc|cccc}
     \hline
     \multirow{2}{*}{Method}  &\multicolumn{4}{c|}{ECSSD} &\multicolumn{4}{c|}{DUT-OMRON} &\multicolumn{4}{c|}{PASCAL-S} &\multicolumn{4}{c|}{HKU-IS} &\multicolumn{4}{c}{DUTS-TE} \\
     \cline{2-21}
     &$S_{\alpha}\uparrow$ &$F_{\beta}\uparrow$ &$E_{\xi}\uparrow$ &$\mathcal{M}\downarrow$ &$S_{\alpha}\uparrow$ &$F_{\beta}\uparrow$ &$E_{\xi}\uparrow$ &$\mathcal{M}\downarrow$ &$S_{\alpha}\uparrow$ &$F_{\beta}\uparrow$ &$E_{\xi}\uparrow$ &$\mathcal{M}\downarrow$ &$S_{\alpha}\uparrow$ &$F_{\beta}\uparrow$ &$E_{\xi}\uparrow$ &$\mathcal{M}\downarrow$ &$S_{\alpha}\uparrow$ &$F_{\beta}\uparrow$ &$E_{\xi}\uparrow$ &$\mathcal{M}\downarrow$ \\
     \hline
     Amulet          &0.893 &0.869 &0.912 &0.059 &0.780 &0.647 &0.784 &0.098 &0.820 &0.768 &0.829 &0.098 &0.884 &0.842 &0.914 &0.051 &0.803 &0.671 &0.798 &0.085 \\
     DSS             &0.882 &0.900 &0.924 &0.053 &0.788 &0.729 &0.846 &0.066 &0.797 &0.804 &0.851 &0.096 &0.880 &0.856 &0.927 &0.050 &0.824 &0.789 &0.885 &0.056 \\
     R$^3$Net        &0.910 &0.914 &0.940 &0.040 &0.817 &0.748 &0.859 &0.062 &0.805 &0.803 &0.845 &0.094 &0.895 &0.894 &0.945 &0.035 &-- &-- &-- &-- \\
     PAGR            &0.889 &0.894 &0.917 &0.061 &0.775 &0.711 &0.843 &0.071 &0.818 &0.807 &0.854 &0.093 &0.889 &0.887 &0.941 &0.047 &0.838 &0.784 &0.883 &0.056 \\
     PiCANet         &0.917 &0.886 &0.927 &0.046 &0.832 &0.717 &0.848 &0.065 &0.854 &0.804 &0.862 &0.076 &0.905 &0.870 &0.941 &0.044 &0.869 &0.759 &0.873 &0.051 \\
     BANet           &0.924 &0.923 &0.953 &0.035 &0.832 &0.746 &0.865 &0.059 &0.852 &0.835 &0.887 &0.070 &0.913 &0.899 &0.955 &0.032 &0.879 &0.815 &0.907 &0.040 \\
     SCRN            &0.927 &0.918 &0.942 &0.037 &0.837 &0.746 &0.869 &0.056 &\textbf{0.868} &0.837 &0.887 &0.064 &0.917 &0.897 &0.954 &0.033 &0.885 &0.808 &0.901 &0.040 \\
     EGNet           &0.925 &0.920 &0.947 &0.037 &0.841 &0.756 &0.874 &0.053 &0.852 &0.828 &0.877 &0.075 &0.918 &0.901 &0.956 &0.031 &0.887 &0.815 &0.907 &0.039 \\
     PAGE            &0.912 &0.906 &0.943 &0.042 &0.824 &0.736 &0.860 &0.062 &0.839 &0.814 &0.878 &0.078 &0.904 &0.884 &0.948 &0.036 &0.854 &0.777 &0.886 &0.052 \\
     AFNet           &0.913 &0.908 &0.941 &0.042 &0.826 &0.739 &0.859 &0.057 &0.849 &0.826 &0.886 &0.071 &0.906 &0.889 &0.949 &0.036 &0.867 &0.792 &0.895 &0.046 \\
     CPD             &0.918 &0.917 &0.949 &0.037 &0.825 &0.747 &0.873 &0.056 &0.847 &0.829 &0.887 &0.072 &0.906 &0.891 &0.952 &0.034 &0.869 &0.805 &0.904 &0.043 \\
     PoolNet         &0.926 &0.919 &0.948 &0.035 &0.831 &0.752 &0.874 &0.054 &0.866 &0.837 &0.888 &0.065 &0.919 &0.903 &0.959 &0.030 &0.886 &0.819 &0.912 &0.037 \\
     CapSal          &0.826 &0.825 &0.866 &0.074 &0.674 &0.564 &0.703 &0.096 &0.838 &0.825 &0.878 &0.073 &0.850 &0.843 &0.907 &0.058 &0.815 &0.755 &0.866 &0.062 \\
     BASNet          &0.916 &0.880 &0.921 &0.037 &0.836 &0.756 &0.869 &0.056 &0.837 &0.779 &0.852 &0.077 &0.908 &0.898 &0.947 &0.033 &0.866 &0.791 &0.884 &0.048 \\
     F$^3$Net        &0.924 &0.925 &0.946 &0.033 &\textbf{0.838} &0.766 &0.876 &0.053 &0.860 &0.845 &0.893 &0.063 &0.917 &0.909 &0.959 &0.028 &0.888 &0.840 &0.918 &0.035 \\
     MINet           &\textbf{0.925} &0.924 &\textbf{0.953} &0.033 &0.833 &0.755 &0.873 &0.055 &0.856 &0.840 &0.898 &0.064 &0.920 &0.908 &0.961 &0.028 &0.884 &0.828 &0.917 &0.037 \\
     LC$^3$Net(ours) &\textbf{0.925} &\textbf{0.928} &\textbf{0.953} &\textbf{0.032} &0.837 &\textbf{0.768} &\textbf{0.878} &\textbf{0.051} &0.866 &\textbf{0.855} &\textbf{0.904} &\textbf{0.059} &\textbf{0.921} &\textbf{0.918} &\textbf{0.963} &\textbf{0.028} &\textbf{0.893} &\textbf{0.854} &\textbf{0.928} &\textbf{0.033} \\
     \hline
  \end{tabular}
  \label{Table 1}
\end{table*}

\begin{figure*} [!t]
  \centering
  \includegraphics[width=1\textwidth]{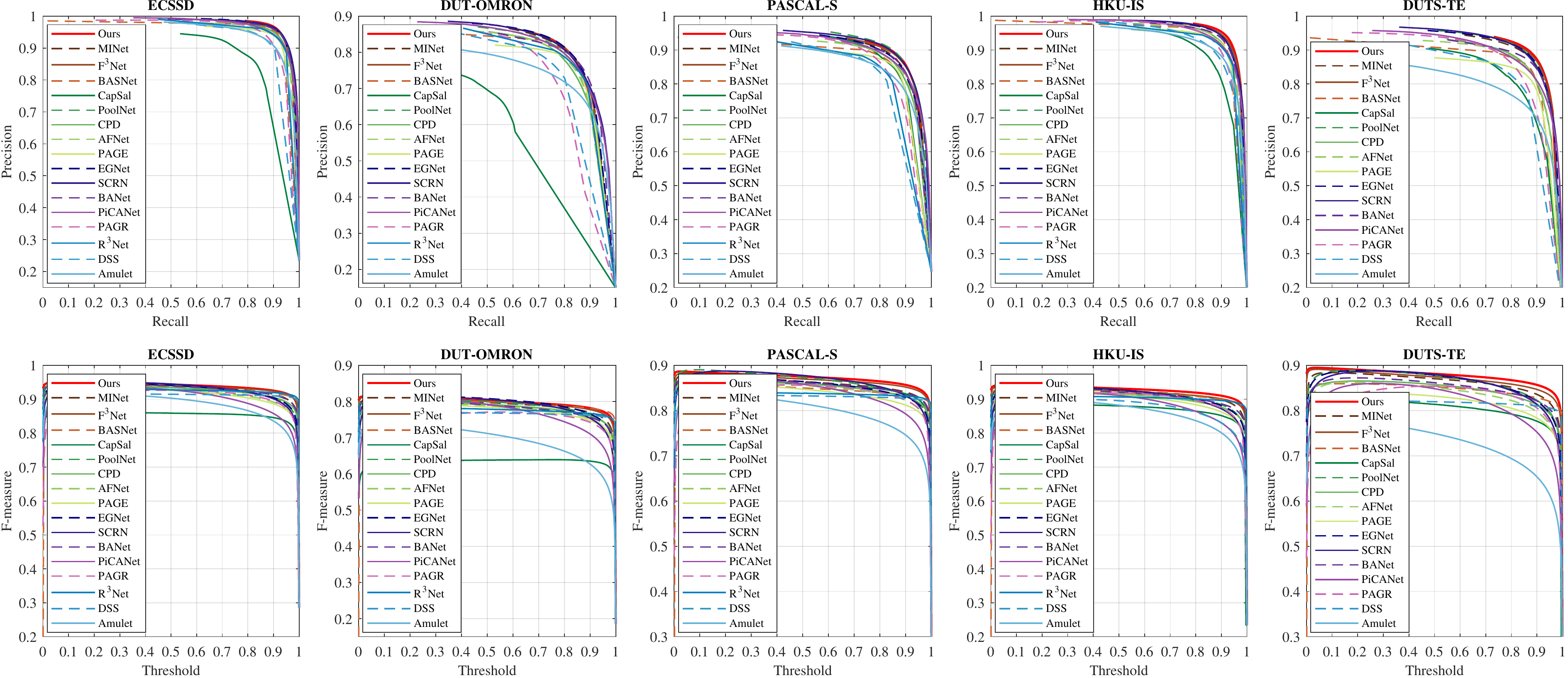}
  \caption{The precision-recall and F-measure curves of different methods on five datasets. The first row shows precision-recall curve and the second row shows F-measure curve.}
  %\vspace{-12pt}
  \label{Figure 6}
\end{figure*}

\subsection{Loss Function}

The binary cross entropy (BCE) loss and intersection over union (IoU) loss are jointly employed for supervision. They enforce local constraint and global constraint, respectively.

BCE loss and IoU loss can be expressed as:
\begin{equation}
  \mathcal{L}_{bce} = -\sum\limits_{p\in P,g\in G} (g\log p + (1-g)\log(1-p)),
\end{equation}
and
\begin{equation}
  \mathcal{L}_{iou} = 1-\frac{\sum\limits_{p\in P,g\in G} (p\times g)}{\sum\limits_{p\in P,g\in G} (p+g-p\times g)},
\end{equation}
where $P=\{p|0<p<1\}$ and $G=\{g|0<p<1\}$ denote the prediction and ground truth, respectively.

In our network, several predictions are generated over different parts of the same layer together with even different layers. There are two categories of losses derived from the dominant streams and auxiliary streams, corresponding to dominant losses and auxiliary losses.

Therefore, dominant losses and auxiliary losses can be computed as:
\begin{equation}
  \mathcal{L}_{dom}^k = \frac{1}{3}\sum\limits_{t=1}^3 (\mathcal{L}_{bce}(P_k^{(t)}, G) + \lambda\mathcal{L}_{iou}(P_k^{(t)}, G)),
\end{equation}
and
\begin{equation}
  \mathcal{L}_{aux}^k = \mathcal{L}_{bce}(P_k, G) + \lambda\mathcal{L}_{iou}(P_k, G),
\end{equation}
where $\lambda$ is the hyperparameter that balances the two terms. For simplicity, it is set to 1.

Finally, the total loss function of the network can be written as:
\begin{equation}
  \mathcal{L}_{total} = \sum\limits_{k=2}^2 \mathcal{L}_{dom}^k + \mu \sum\limits_{k=3}^5 \eta_k \mathcal{L}_{aux}^k,
\end{equation}
where $\mu$ is a trade-off parameter set to 1 empirically, $\eta_k$ $(k=3,4,5)$ are the weight coefficients set to 1/2, 1/4 and 1/8, respectively.

\begin{figure*} [t]
  \centering
  \includegraphics[width=1\textwidth]{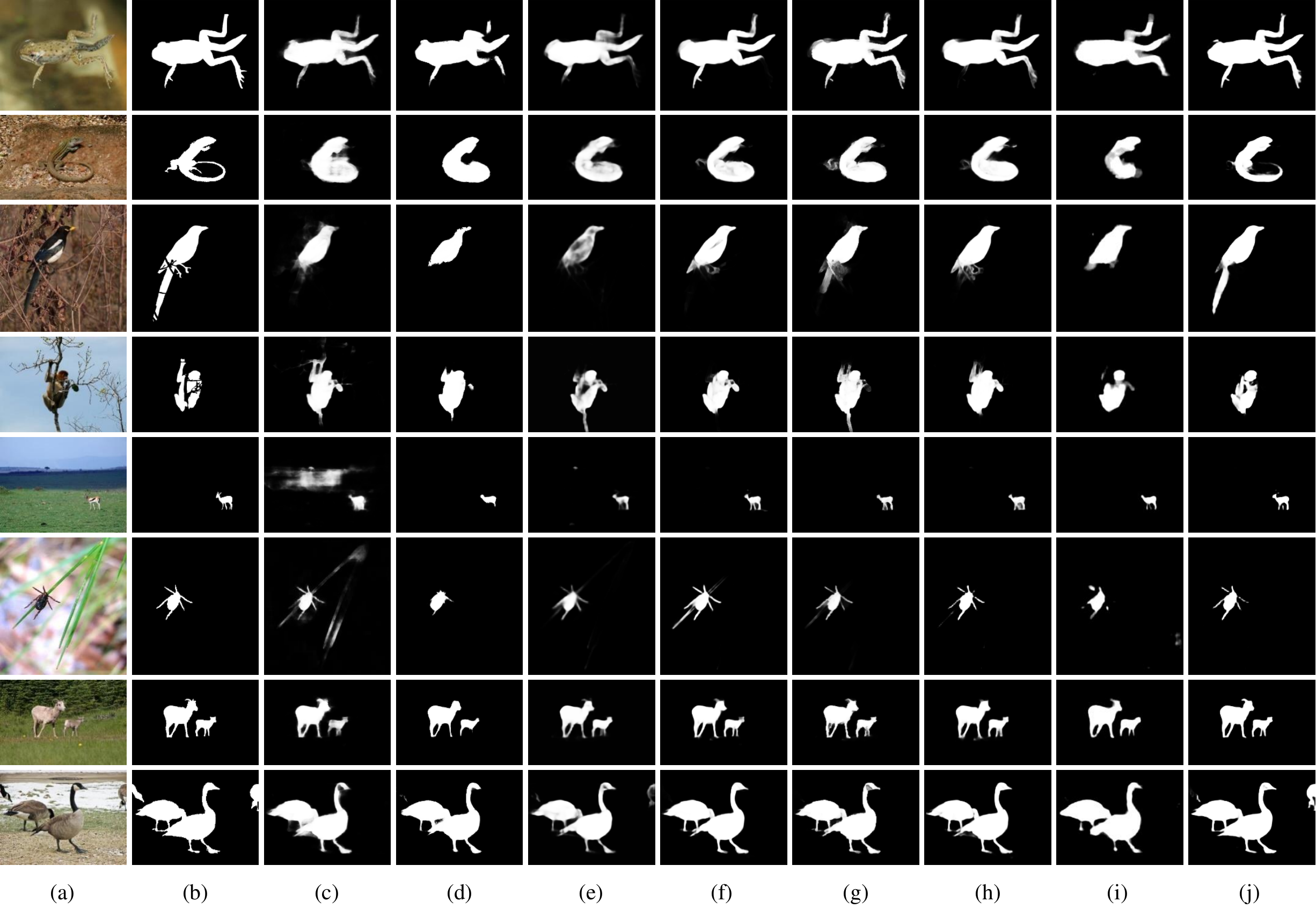}
  \caption{Qualitative comparison of different methods in some representative scenes. (a) Image; (b) Ground truth; (c) Amulet; (d) DSS; (e) RAGR; (f) EGNet; (g) AFNet; (h) PoolNet; (i) CapSal; (j) Ours.}
  %\vspace{-12pt}
  \label{Figure 7}
\end{figure*}

\section{Experiments}
\label{Section 4}

\subsection{Experimental Setup}

\noindent \textbf{Datasets.} We choose five popular datasets, which are ECSSD \cite{Yan2013Hierarchical}, DUT-OMRON \cite{Yang2013Saliency}, PASCAL-S \cite{Li2014The}, HKU-IS \cite{Li2015Visual} and DUTS \cite{Wang2017Learning}, with 1000, 5168, 850, 4447 and 15572 images and their annotations, respectively. DUT-TR, the training set of DUTS, has 10553 images and corresponding annotated maps, while DUT-TE, the testing set of DUTS, has 5019 images and corresponding annotated maps. Among them, DUT-TR is used for training, while the rest are used for testing. Note that only 1447 images and maps in HKU-IS are used following \cite{Li2015Visual, Hou2017Deeply, Chen2018Reverse}.

\noindent \textbf{Evaluation Metrics.} We select four typical evaluation metrics, which are S-measure ($S_{\alpha}$) \cite{Fan2017Structure-measure}, average F-measure ($F_{\beta}$) \cite{Achanta2009Frequency-tuned}, E-measure ($E_{\xi}$) \cite{Fan2018Enhanced-alignment} and mean absolute error ($\mathcal{M}$) \cite{Perazzi2012Saliency}. They are defined as:
\begin{equation}
  S_{\alpha} = \frac{S_o+S_r}{2},
\end{equation}
where $S_o$ and $S_r$ assess object-aware and region-aware structure similarities,
\begin{equation}
  F_{\beta} = \frac{(1+\beta^{2})\times Precision\times Recall}{\beta^{2}\times Precision+Recall},
\end{equation}
where $\beta^{2}$ is set as 0.3 to emphasize more on the precision than recall suggested in \cite{Borji2015Salient},
\begin{equation}
  E_{\xi} = \frac{1}{W\times H} \sum\limits_{x=1}^W\sum\limits_{y=1}^H\phi(x,y),
\end{equation}
where $W$ and $H$ are width and height, and $\phi$ is the enhanced alignment matrix,
\begin{equation}
  \mathcal{M} = \frac{1}{W\times H} \sum\limits_{x=1}^W\sum\limits_{y=1}^H|P(x,y)-G(x,y)|,
\end{equation}
where $P$ and $G$ denote the prediction and ground truth. Additionally, we plot the precision-recall curve and F-measure curve.

\noindent \textbf{Implementation Details.}
All experiments are implemented on a GeForce RTX 2080 Ti GPU using PyTorch repository. Data augmentation technology adopts horizontal flipping, random cropping and multi-scale input images. ResNet-50 \cite{He2016Deep}, pre-trained on ImageNet \cite{Deng2009ImageNet}, is employed as the backbone network. The model is optimized by stochastic gradient descent with batch size of 32, momentum of 0.9 and weight decay of 5e-4. Warm-up and decay strategies are utilized to adjust the learning rate under the maximum learning rate of 0.05. The epoch is set to 50. During inference, each image is resized to 352$\times$352. Any
%pre-processing and
post-processing operations (e.g., CRF \cite{Krahenbuhl2011Efficient}) is not applied.

\subsection{Comparison with State-of-the-arts}

We compare the proposed method with 16 state-of-the-art methods, including Amulet \cite{Zhang2017Amulet}, DSS \cite{Hou2017Deeply}, R$^3$Net \cite{Deng2018R^3Net}, PAGR \cite{Zhang2018Progressive}, PiCANet \cite{Liu2018PiCANet}, BANet \cite{Su2019Selectivity}, SCRN \cite{Wu2019Stacked}, EGNet \cite{Zhao2019EGNet}, PAGE \cite{Wang2019Salient}, AFNet \cite{Feng2019Attentive}, CPD \cite{Wu2019Cascaded}, PoolNet \cite{Liu2019A}, CapSal \cite{Zhang2019CapSal}, BASNet \cite{Qin2019BASNet}, F$^3$Net \cite{Wei2020F^3Net} and MINet \cite{Pang2020Multi-scale}. For fair comparison, the saliency maps of competing methods are provided by authors or generated by authorized codes. If the competitors has multiple versions based on different backbone networks (e.g., VGG-16 \cite{Simonyan2014Very}, ResNeXt-101 \cite{Xie2017Aggregated}), we give priority to using the same version as ours.

\begin{table} [t]
  \centering
  \caption{Impact of each component. 'B'$=$FCB, 'M'$=$DCM, 'D'$=$BCD.}
  \setlength{\tabcolsep}{2pt}
  \renewcommand\arraystretch{1.1}
  \small
  \begin{tabular}{c|cccc|cccc}
    \hline
    &\multicolumn{4}{c|}{DUT-OMRON} &\multicolumn{4}{c}{DUTS-TE} \\
    \cline{2-9}
    &$S_{\alpha}\uparrow$ &$F_{\beta}\uparrow$ &$E_{\xi}\uparrow$ &$\mathcal{M}\downarrow$ &$S_{\alpha}\uparrow$ &$F_{\beta}\uparrow$ &$E_{\xi}\uparrow$ &$\mathcal{M}\downarrow$ \\
    \hline
    Baseline     &0.784 &0.725 &0.843 &0.069 &0.827 &0.792 &0.887 &0.050 \\
    \hline
    $+$B         &0.821 &0.719 &0.850 &0.064 &0.874 &0.819 &0.903 &0.046 \\
    $+$B$+$M     &0.831 &0.758 &0.868 &0.058 &0.886 &0.839 &0.917 &0.039 \\
    $+$B$+$M$+$D &\textbf{0.837} &\textbf{0.768} &\textbf{0.878} &\textbf{0.051} &\textbf{0.893} &\textbf{0.854} &\textbf{0.928} &\textbf{0.033} \\
    \hline
  \end{tabular}
  \label{Table 2}
\end{table}

\noindent \textbf{Quantitative Comparison.}
The results of quantitative comparison are listed in Table \ref{Table 1}. As can be seen, the performance of our method is outstanding on all datasets, especially on ECSSD, HKU-IS and DUTS-TE, in terms of all evaluation metrics. Notably, our method reaches 0.059 on PASCAL-S in terms of $\mathcal{M}$, which surpasses other methods by a large margin. In addition, the precision-recall and F-measure curves are shown in Figure \ref{Figure 6}. It can be seen that the curves of our method are more stable and higher than those of others.

\noindent \textbf{Qualitative Comparison.}
The results of qualitative comparison are shown in Figure \ref{Figure 7}. A variety of challenging scenarios are picked, in which the salient objects vary from the contrast, texture, size, number and so on. Compared to other methods, our method can attain the excellent performance. It is obvious that our method is capable of handling these complex situations accurately, covering low contrast (i.e., 1$^{st}$ and 2$^{nd}$ rows), cluttered background (i.e., 3$^{rd}$ and 4$^{th}$ rows) small object (i.e., 5$^{th}$ and 6$^{th}$ rows) and multiple object (i.e., 7$^{th}$ and 8$^{th}$ rows). In short, these visualization results strongly verify the good robustness of our method to the various salient objects.

\subsection{Ablation Study}

We study the impact of each component and sub-component in the proposed method.

\noindent \textbf{Impact of Each Component.}
The results of impact of each component are listed in Table \ref{Table 2}. The baseline refers to the network like feature pyramid networks (FPNs) \cite{Lin2017Feature}. With the embedding of the increasing components, we can find that the performance of our method is continuously improved, and the optimal performance is obtained by using all components, which reveals that all the components are necessary to realize this method. Moreover, comparing our method with the baseline, we obtain gains of 5.3\% for $S_{\alpha}$, 4.3\% for $F_{\beta}$, 3.5\% for $E_{\xi}$ and 1.8\% for $\mathcal{M}$ on DUT-OMRON, and gains of 6.6\% for $S_{\alpha}$, 6.2\% for $F_{\beta}$, 4.1\% for $E_{\xi}$ and 1.7\% for $\mathcal{M}$ on DUT-TE, respectively.

\noindent \textbf{Impact of Each Sub-component.}
The results of impact of DCM-U and DCM-D are listed in Table \ref{Table 3}. We can observe that 'w$/$ M' performs best, followed by 'w$/$o M-U' and 'w$/$o M-D', and 'w$/$o M' performs worst. Moreover, 'w$/$o M-U' has a slight enhancement over 'w$/$o M', whereas 'w$/$o M-D' has a evident enhancement over 'w$/$o M', indicating the overwhelming significance of DCM-U than DCM-D, despite both are vital.
The results of impact of BCD$_1$, BCD$_2$ and BCD$_3$ are listed in Table \ref{Table 4}. We can observe that our method gets the most satisfactory results when '$\Sigma$$=$3'.
Moreover, the enhancement by '$\Sigma$$=$1', '$\Sigma$$=$2' and '$\Sigma$$=$3' is steadily incremental, indicating the significance of the trapezoidal network of BCD$_1$, BCD$_2$ and BCD$_3$.

\begin{table} [t]
  \centering
  \caption{Impact of DCM-U and DCM-D. 'M-U'$=$DCM-U, 'M-D'$=$DCM-D,
   'w$/$ $\star$' and 'w$/$o $\star$' mean with$/$without $\star$.}
  \setlength{\tabcolsep}{2pt}
  \renewcommand\arraystretch{1.1}
  \small
  \begin{tabular}{c|cccc|cccc}
    \hline
    &\multicolumn{4}{c|}{DUT-OMRON} &\multicolumn{4}{c}{DUTS-TE} \\
    \cline{2-9}
    &$S_{\alpha}\uparrow$ &$F_{\beta}\uparrow$ &$E_{\xi}\uparrow$ &$\mathcal{M}\downarrow$ &$S_{\alpha}\uparrow$ &$F_{\beta}\uparrow$ &$E_{\xi}\uparrow$ &$\mathcal{M}\downarrow$ \\
    \hline
    w$/$o M   &0.784 &0.725 &0.843 &0.069 &0.827 &0.792 &0.887 &0.050 \\
    w$/$o M-U &0.803 &0.735 &0.849 &0.065 &0.843 &0.833 &0.899 &0.053 \\
    w$/$o M-D &0.823 &0.757 &0.858 &0.054 &0.884 &0.840 &0.912 &0.036 \\
    w$/$ M    &\textbf{0.837} &\textbf{0.768} &\textbf{0.878} &\textbf{0.051} &\textbf{0.893} &\textbf{0.854} &\textbf{0.928} &\textbf{0.033} \\
    \hline
  \end{tabular}
  \label{Table 3}
\end{table}

\begin{table} [t]
  \centering
  \caption{Impact of BCD$_1$, BCD$_2$ and BCD$_3$. '$\Sigma$=1,2,3' mean BCD$_1$, BCD$_1$$+$BCD$_2$, BCD$_1$$+$BCD$_2$$+$BCD$_3$.}
  \setlength{\tabcolsep}{2pt}
  \renewcommand\arraystretch{1.1}
  \small
  \begin{tabular}{c|cccc|cccc}
    \hline
    &\multicolumn{4}{c|}{DUT-OMRON} &\multicolumn{4}{c}{DUTS-TE} \\
    \cline{2-9}
    &$S_{\alpha}\uparrow$ &$F_{\beta}\uparrow$ &$E_{\xi}\uparrow$ &$\mathcal{M}\downarrow$ &$S_{\alpha}\uparrow$ &$F_{\beta}\uparrow$ &$E_{\xi}\uparrow$ &$\mathcal{M}\downarrow$ \\
    \hline
    $\Sigma$$=$1 &0.832          &0.763          &0.875          &0.054          &0.886          &0.839          &0.916          &0.037 \\
    $\Sigma$$=$2 &0.836          &0.765          &0.873          &0.052          &0.888          &0.845          &0.922          &0.035 \\
    $\Sigma$$=$3 &\textbf{0.837} &\textbf{0.768} &\textbf{0.878} &\textbf{0.051} &\textbf{0.893} &\textbf{0.854} &\textbf{0.928} &\textbf{0.033} \\
    \hline
  \end{tabular}
  \label{Table 4}
\end{table}

\section{Conclusion}
\label{Section 5}

In this paper, we propose a ladder context correlation complementary network (LC$^3$Net) for precise salient object detection. Different from the existing methods that ignore the gap between high level and low level features, we aim to explore the inherent complementarity of correlated contextual information. To assist the automatic collection of information on the diversity of initial features, we propose a filterable convolution block (FCB). To facilitate the intimate aggregation of different levels of features, we propose a dense cross module (DCM). To help the progressive shrinkage of multi-scale features from coarse to fine, we propose a bidirectional compression decoder (BCD). These three components of our method play a role in achieving remarkable detection results. Comprehensive experiments quantitatively and qualitatively demonstrate that the proposed LC$^3$Net can outperform 16 state-of-the-arts.

\section*{CRediT authorship contribution statement}

\textbf{Xian Fang:}
Conceptualization, Methodology, Validation, Formal analysis, Investigation, Writing - original draft, Visualization.
\textbf{Jinchao Zhu}:
Data Curation, Writing - review \& editing, Visualization.
\textbf{Xiuli Shao}:
Writing - review \& editing.
\textbf{Hongpeng Wang}:
Writing - review \& editing, Funding acquisition.

\section*{Declaration of competing interest}

The authors declare that they have no known competing financial interests or personal relationships that could have appeared to influence the work reported in this paper.

\section*{Acknowledgement}

This research was supported by the National Key R\&D Program of China under Grant 2019YFB1311804, the National Natural Science Foundation of China under Grant 61973173, 91848108 and 91848203, and the Technology Research and Development Program of Tianjin under Grant 18ZXZNGX00340 and 20YFZCSY00830.

\section*{References}

\bibliography{BibTex}

\end{document}